\documentclass[10pt,twocolumn,letterpaper]{article}

\usepackage{iccv}
\usepackage{times}
\usepackage{epsfig}
\usepackage{graphicx}
\usepackage{amsmath}
\usepackage{amssymb}
\usepackage{subcaption}
\usepackage{makecell}
\usepackage{multirow}
\usepackage[export]{adjustbox}
\usepackage{mwe}
\usepackage{enumitem}
\usepackage{makecell}
\usepackage[numbers]{natbib}
\usepackage{multirow}
\DeclareMathOperator*{\argmax}{argmax}

\usepackage[pagebackref=true,breaklinks=true,letterpaper=true,colorlinks,bookmarks=false]{hyperref}

\iccvfinalcopy 

\def\iccvPaperID{6} 

\ificcvfinal\fi
\begin{document}

\title{Deep Metric Transfer for Label Propagation with Limited Annotated Data}
\renewcommand\footnotemark{}
\author{
	Bin Liu$^{1\dagger}$~~~~~~~~~~Zhirong Wu$^{2\dagger}$~~~~~~~~~~Han Hu$^{2*}$\thanks{$^\dagger$Equal contribution. $^\star$Corresponding author. The work is done when Bin Liu is an intern at MSRA.}~~~~~~~~Stephen Lin$^{2}$ \\
	$^1$Tsinghua University~~~~~$^2$Microsoft Research Asia \\
	{\tt\small liub18@mails.tsinghua.edu.cn} ~~
	{\tt\small \{wuzhiron,hanhu,stevelin\}@microsoft.com}}
	
\maketitle

\begin{abstract}
We study object recognition under the constraint that each object class is only represented by very few observations.
Semi-supervised learning, transfer learning, and few-shot recognition all concern with achieving fast generalization with few labeled data.
In this paper, we propose a generic framework that utilize unlabeled data to aid generalization for all three tasks.
Our approach is to create much more training data through label propagation from the few labeled examples to a vast collection of unannotated images.
The main contribution of the paper is that we show such a label propagation scheme can be highly effective when the similarity metric used for propagation is transferred from other related domains.
We test various combinations of supervised and unsupervised metric learning methods with various label propagation algorithms. 
We find that our framework is very generic without being
sensitive to any specific techniques.
By taking advantage of unlabeled data in this way, we achieve significant improvements on all three tasks.
Code is availble at \url{http://github.com/Microsoft/metric-transfer.pytorch}.
\end{abstract}

\section{Introduction}

\begin{figure}[t]
	\centering
	\includegraphics[width=1.0\linewidth]{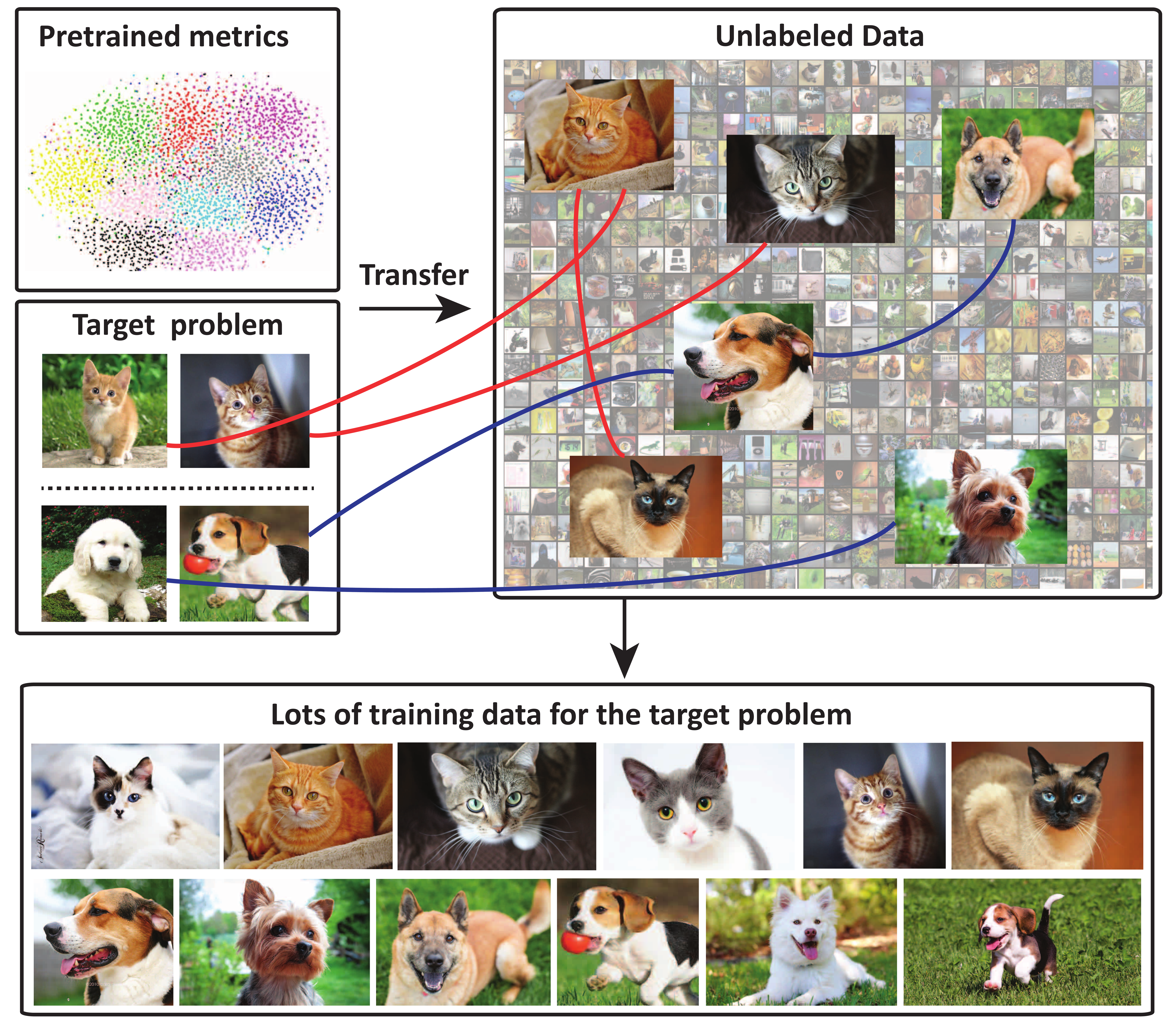}
	\caption{Overview of the approach. Often, object categories are represented by very few images. We transfer a metric learned from another domain and propagate the labels from the few labeled images to a vast collection of unannotated images. We show this can reliably create much more labeled data for the target problem.
	}
	\label{fig:teaser}
\end{figure}

We address the problem of object recognition from a very small amount of labeled data.
This problem is of particular importance when limited labels can be collected due to either time or financial constraints.
Though this is a difficult challenge, we are encouraged by evidence from cognitive science suggesting that infants can quickly learn new concepts from very few examples~\cite{lake2011one,carey1978acquiring}.

Many recognition problems in computer vision are concerned with learning on few labeled data.
Semi-supervised learning, transfer learning, and few-shot recognition 
all aim to achieve fast generalization from few examples,
by leveraging unlabeled data or labeled data from other domains.

The fundamental difficulty of this problem is that naive supervised training with very few examples results in severe over-fitting.
Because of this, prior work in semi-supervised learning rely on strong regularizations
such as augmentations~\cite{gastaldi2017shake}, temporal consistency~\cite{laine2016temporal}, and adversarial examples~\cite{miyato2017virtual} to improve performance.
Some related works in few-shot learning do not even refine an online classifier.
Instead, they simply apply the similarity metric learned from training categories
to new categories without adaptation.
Meta-learning~\cite{finn2017model} seeks to optimize an online parametric classifier with few samples, but
under the assumption that just a few steps of optimization will lead to effective
generalization with less overfitting.
These approaches indirectly address the inherent problem of limited training data.

In this paper, we propose a new framework of label propagation via metric transfer to tackle the problem of limited training data.
We propagate labels to an unlabeled dataset, so that training a supervised model with great learning capacity no longer faces over-fitting.
This approach is related to work on ``pseudo-labeling"~\cite{lee2013pseudo,rosenberg2005semi}, where the model is bootstrapped from limited data and trained on the new data/label pairs it infers.
However, that is unlikely to work well when the labeled data is scarce, since the initial model is likely to be poor.
Instead of bootstrapping, our work transfers the metric learned from another related domain, and thus provides a much better generalization ability.

 Our approach works with three data domains: a source domain to learn a similarity metric, few labeled examples to define the target problem, and an unlabeled dataset in which to propagate labels.
 As in Figure~\ref{fig:teaser}, we first learn a similarity metric on the source domain, which can be either labeled or unlabeled.
 Supervised learning or unsupervised (self-supervised) learning is used to learn the metric accordingly.
 Then, given few observations of the target problem, we propagate the labels from these observations to the unlabeled dataset using the metric learned in the source domain. 
 This creates an abundance of labeled data for learning a classifier.
 Finally, we train a standard supervised model using the propagated labels.
 
 The main contribution of this work is the metric transfer approach for label propagation.
 By studying different combinations of metric pretraining methods (\textit{e.g.} unsupervised, supervised) and label propagation algorithms (\textit{e.g.} nearest neighbors, spectral clustering), we find that our metric transfer approach on unlabeled data is general enough to work effectively for many settings.
 For semi-supervised learning on CIFAR10 and ImageNet, we obtain an absolute $20\%$ improvement over the state-of-the-art when labeled data is limited ($5-10$ labels per category).
 We also achieve a $10\%$ improvement on transferring representations from ImageNet to CIFAR10 for transfer learning,
 and $6\%$ improved performance for few-shot recognition
 on the mini-ImageNet benchmark.

Due to this generic framework, our work also brings individual insights into the respective tasks we studied: 
1) for semi-supervised learning, algorithms may better develop from unsupervised learning, as opposed to using unlabeled data for regularization. 
2) for transfer learning, we propose an alternative method for transferring knowledge other than the dominant finetuning approach.
3) for few-shot recognition, in certain scenarios, unlabeled data in the target domain is more beneficial than labeled data in the source domain.

\section{Related Work}

\vspace{3pt}

\noindent \textbf{Large-scale Recognition.}
To solve a computer vision problem,
it has become a common practice to build a large-scale dataset~\cite{everingham2010pascal,deng2009imagenet} and train deep neural networks~\cite{krizhevsky2012imagenet,simonyan2014very} on it.
This philosophy has achieved unprecedented success on
many important computer vision problems~\cite{everingham2010pascal,lin2014microsoft,russakovsky2015imagenet}.
However, constructing a large-scale dataset is often time-consuming and expensive, and this has motivated work on unsupervised learning and problems defined on few labeled samples.

\vspace{3pt}

\noindent \textbf{Semi-supervised Learning.}
Semi-supervised learning~\cite{weston2012deep} is a problem that lies in between supervised learning and unsupervised learning. 
It aims to make more accurate predictions by leveraging a large amount of unlabeled data than 
by relying on the labeled data alone.
In the era of deep learning, 
one line of work leverages unlabeled data through deep generative models~\cite{kingma2014semi,rasmus2015semi}.
However, training of generative models is often unstable, making it
tricky to work with recognition tasks.
Recent efforts on semi-supervised learning focus on regularization by
self-ensembling through consistency loss, such as temporal ensembling~\cite{laine2016temporal}, adversarial ensembling~\cite{miyato2017virtual}, teacher-student distillation~\cite{tarvainen2017mean}, and cross-view ensembling~\cite{clark_cross-view_2018}.  
The pseudo-labeling approach~\cite{lee2013pseudo,rosenberg2005semi} initializes a model on a smalled labeled dataset and bootstraps on the new data it predicts.
This tends to fail when the labeled set is small.

Our work is most closely related to the transductive  approaches~\cite{joachims2003transductive, zhou2004learning}.
Prior work~\cite{fergus2009semi} in computer vision shows that label propagation can work well with handcrafted GIST descriptors.
We bring it to the context of deep learning, and
demonstrate that metric transfer may further 
improve the accuracy of label propagation.
\vspace{3pt}

\noindent \textbf{Few-shot Recognition.}
Given some training data in training categories, few-shot recognition~\cite{carey1978acquiring} requires the classifier to generalize to new categories from observing very few examples, often 1-shot or 5-shot.
A body of work approaches this problem by offline metric learning~\cite{vinyals2016matching,snell2017prototypical,wu2018improving}, where a generic similarity metric is learned on the training data and directly transferred to the new categories using simple nearest neighbor classifiers without further adaptation.
Recent works on meta-learning~\cite{finn2017model,li2017meta,mishra2018simple} take a learning-to-learn approach using online algorithms. In order not to overfit to the few examples, they develop meta-learners to find a common embedding space, which can be further finetuned with fast convergence to the target problem.
Recent works~\cite{ren2018meta,garcia2017few} using meta-learning consider the combined problem of semi-supervised learning and few-shot recognition, by allowing access to unlabeled data in few-shot recognition.
This drives few-shot recognition into more realistic scenarios.
We follow this setting as we study few-shot recognition.

\vspace{3pt}

\noindent \textbf{Transfer Learning.}
Since the inception of the ImageNet challenge~\cite{russakovsky2015imagenet}, transfer learning 
has emerged almost everywhere in visual recognition, such as in object detection~\cite{girshick2014rich} and semantic segmentation~\cite{long2015fully}, by simply
transferring the network weights learned on ImageNet classification and finetuning on the target task.
When the pretraining task and the target task are closely related, 
this tends to generalize much better than training from scratch on the target task alone.
Domain adaptation seeks to address a much more difficult scenario where there is a large gap between the inputs of the source and target domains~\cite{hoffman2017cycada}, for example, between real images and synthetic images.
What we study in this paper is metric transfer.
Different from prior work~\cite{xu2017unified} that employ metric transfer just to reduce the distribution divergence of different domains, we use metric transfer to propagate labels.
Through this, we show that metric propagation is an effective method for
learning with small data.

\section{Approach}

To deal with the shortage of labeled data, our approach is to enlarge it
by propagating labels from annotated images to unlabeled data using the similarity
metric between data pairs.
The creation of much more labeled data enables us to train deep neural networks
to their full learning capacity.

Our framework works on three data domains: the source domain $\mathcal{S}$, the target domain $\mathcal{T}$, and additional unlabeled data $\mathcal{U}$. The source domain $\mathcal{S}$ can be labeled or unlabeled with abundant data, and it is used to learn a generic similarity metric between data pairs. The target domain $\mathcal{T}$ only has few labeled data, but it defines the problem we want to optimize. The unlabeled data $\mathcal{U}$ is the resource in which to propagate labels, and may potentially contain similar classes to the task defined in $\mathcal{T}$. It may or may not have overlapping classes with $\mathcal{S}$.

The approach we propose in this paper is very general, 
suggesting that a spectrum of metric pretraining and label propagation algorithms can all work well in this framework.
Below we introduce our method in details, and overview several metric learning and label propagation methods we used for our experiments.

\subsection{Metric Pretraining}

The source domain $\mathcal{S}$ is used for pretraining a similarity metric between data pairs. 
Ideally, we desire the metric to capture the inherent structure in the target domain $\mathcal{T}$, so that transferring labels from $\mathcal{T}$ is reliable and useful.
For this to happen, we usually hold some prior knowledge about the source $\mathcal{S}$ and the target $\mathcal{T}$. For example, the source domain is sampled from the same distribution as the target domain, but is completely unannotated, or the source domain is annotated with a different task but is closely related to the target.
Formally, a similarity metric $s_{ij}$ between data $x_i$ and $x_j$ can be defined as
\begin{equation}
s_{ij} = f(x_i, x_j),
\end{equation}
where $f$ is the similarity function to be learned.
In this work, we use deep neural networks as a parametric model of this similarity function.
The metric can be trained with either supervised or unsupervised methods, depending on whether labels are given in the source domain $\mathcal{S}$.
We briefly review the training algorithms as follows.

\vspace{4pt}
\noindent \textbf{Unsupervised Metric Pretraining} \\
Recently, there has been growing interest in unsupervised learning and self-supervised learning.
Different algorithms are based on different data properties (e.g. color~\cite{zhang2016colorful}, context~\cite{doersch2015unsupervised}, motion~\cite{zhou2017unsupervised}) and thus may vary in performance on the target task we may want to transfer.
However, it is not our intent to give a comprehensive comparison over various methods and choose the best one.
Instead, we show that general unsupervised transfer
is beneficial for label propagation and leads to improved performance. 

In this work, we utilize two unsupervised learning methods: instance discrimination~\cite{wu2018unsupervised} and colorization~\cite{zhang2016colorful}.
For instance discrimination, we treat each instance as a class, and maximize the probability of each example belonging to the class of itself,
\begin{equation}
    P(i|x_i) = \frac{\text{exp} (s_{ii})}{\sum_{j=1}^{n}\text{exp} (s_{ij})}.
\end{equation}

For colorization, the idea is to learn a mapping from grayscale images to colorful ones. Following the original paper~\cite{zhang2016colorful}, instead of predicting raw pixel colors, we quantize the color space into soft bins $q$, and use the cross-entropy loss on the soft bins,
\begin{equation}
\mathcal{L}_{\text{color}} = - \sum_{h,w} q_{h,w} \text{log }(q_{h,w}),
\end{equation}
where $h,w$ are spatial indices. We follow previous work~\cite{doersch2017multi} for applying ResNet to colorization, where we use a base network to map inputs to features, and a head network of three convolutional layers to convert features to colors.
Since colorization does not automatically output a metric, we use the Euclidean distance on the features from the base network to measure similarity.


\vspace{4pt}
\noindent \textbf{Supervised Metric Pretraining} \\
In some scenarios, we have access to a labeled dataset, such as PASCAL VOC and ImageNet, having
commonalities with the target task.
Traditional metric learning with supervision minimizes the
intra-class distance and maximizes the inter-class distance of the labeled samples.
For this purpose, many types of loss functions such as contrastive loss, triplet loss~\cite{hoffer2015deep}, and neighborhood analysis~\cite{goldberger2005neighbourhood} have been proposed.
In this work,
we use neighborhood analysis~\cite{goldberger2005neighbourhood} to learn our metric. Concretely,
we maximize the likelihood of each example being supported
by other examples belonging to the same category,
\begin{equation}
    P(y_i|x_i) =  \frac{\sum_{y_k = y_i}\text{exp} (s_{ik})}{\sum_{j=1}^{n}\text{exp} (s_{ij})}.
\end{equation}

\subsection{Label Propagation}

Given a target $\mathcal{T}$ represented by a small number of labeled examples, and a unlabeled set $\mathcal{U}$, we propagate labels from $\mathcal{T}$ to $\mathcal{U}$ using the similarity function $f(\cdot)$ learned from $\mathcal{S}$.
Suppose $\mathcal{T} = \{(x_1, y_1), (x_2, y_2), ..., (x_{n_t}, y_{n_t}) \}$, and $\mathcal{U} = \{x_{n_t+1}, x_{n_t+2}, ..., x_{n_t+n_u} \}$, where $n_t$ and $n_u$ are the number of images in $\mathcal{T}$, $\mathcal{U}$ respectively. Label $y_i$ is represented as a vector with the ground-truth class element set to $1$ and the others set to $-1$. We consider two propagation algorithms.

\vspace{4pt}
\noindent \textbf{Naive Nearest Neighbors} \\
A straightforward propagation approach is to vote for the class of an unlabeled sample based on its similarity to each of the exemplars in the target set $\mathcal{T}$. For an unlabeled example $x_u \in \mathcal{U}$, we calculate its logits $z_{u,c}$ for every class $c$,
\begin{equation}
z_{u,c} = \frac{1}{n_{t,c}} \sum_{i=1}^{n_t} I(y_{i,c} = 1) \cdot W_{i,u},
\label{eq:knn_propagation}
\end{equation}
where $I(\cdot)$ is the indicator function, $W_{i,u} = \text{exp} \left( f(x_i, x_u) \right))$ denotes the similarity between example $i$ and $u$, and $n_{t,c}$ is the number of labeled images available for class $c$.

The nearest neighbor propagation method is essentially a one-step random walk
where the similarity metric acts as the transition matrix and the indicator function acts as the initial distribution.
The effectiveness of such one-step propagation depends heavily on the quality of the similarity metric.

\begin{figure}[t]
	\centering
	\includegraphics[width=0.47\linewidth]{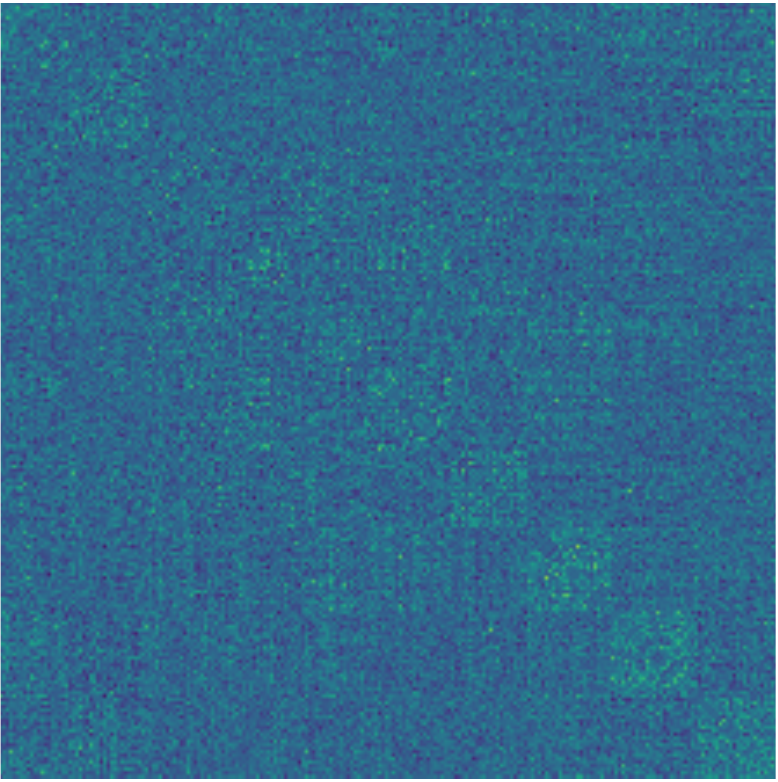}
	\includegraphics[width=0.47\linewidth]{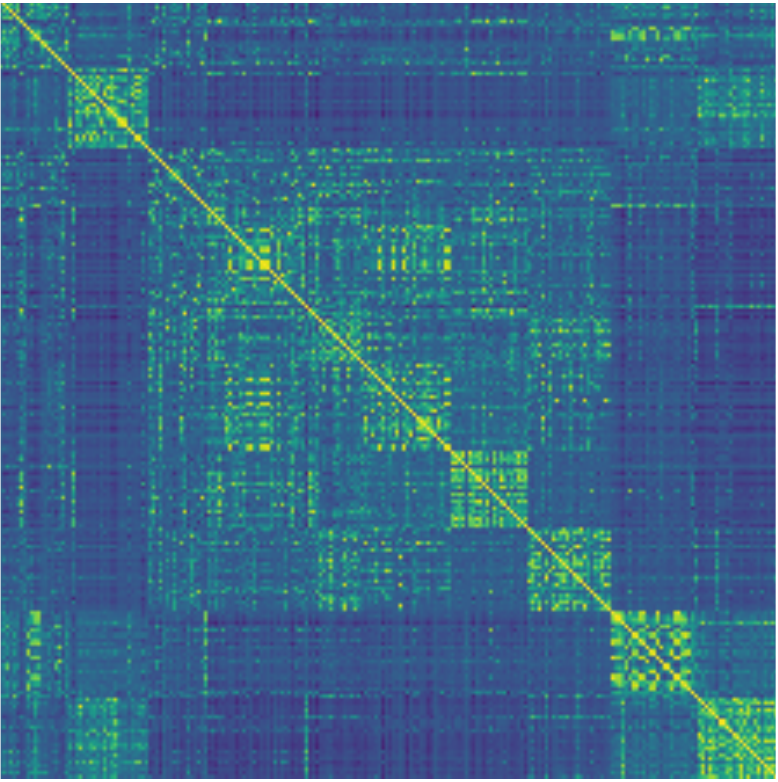}
	\caption{Left: raw similarity matrix. Right: similarity matrix by spectral embedding. Through spectral embedding, sparse similarities are propagated to distant areas to reveal \emph{global} structure. Samples are sorted by their class id for better visualization.}
	\label{fig:propagation}
\end{figure}

In general, it is hard to learn such a metric well, especially when limited supervision is available,
because of the visual diversity of images.
Figure \ref{fig:propagation} (left) shows a typical similarity matrix computed from unsupervised features.
Data points in the similarity matrix are sparsely connected,
thus limiting the one-step label propagation approach.

\vspace{4pt}
\noindent \textbf{Constrained Spectral Clustering} \\
Constrained spectral clustering~\cite{hu2013multi,fergus2009semi} may potentially relieve such a problem. Instead of propagating labels by one step as in the naive nearest neighbor approach, constrained spectral clustering propagates labels through multiple steps by taking advantage of structure within the unlabeled dataset. 
It computes a spectral 
embedding~\cite{shi2000normalized,von2007tutorial} from the original similarity metric, which is then used as the new metric for label propagation.
The spectral embedding is formulated as
\begin{equation}
W' = \sum_{j=2}^{\eta} \frac{1}{\lambda_j} e_j e^T_j,
\end{equation}
where $\lambda_j$ and $e_j$ are the eigenvalues and eigenvectors of the normalized Laplacian in ascending order. 
The Laplacian matrix $L_{\text{sym}}$ is derived from the original similarity metric as
$L_{\text{sym}} = I - D^{-1/2}WD^{-1/2}$, with degree matrix $D =\text{diag}(d)$ and $d_i=\sum_j W_{ij}$. Parameter ${\eta}$ is the total number of eigen components used.

Due to its globalized nature, spectral clustering is able to pass messages between distant areas, which is in contrast to the \emph{local} behavior of the naive nearest neighbors approach.
The embedded metric is usually densely connected and better aligned with object classes, as illustrated in Figure~\ref{fig:propagation} (right). 
Using the same voting approach as in Eqn~(\ref{eq:knn_propagation}), 
labeled propagation can be more accurate than using the original raw similarity metric.

Constrained spectral clustering is also efficient. By following the common practice of using $k$-nearest neighbors to build the similarity graph~\cite{von2007tutorial}, propagating labels to $50k$ images takes about 10 seconds on a regular GPU.

 \begin{table*}[t]
	\setlength{\tabcolsep}{7pt}
	\caption{Ablation study of the mean average precision (mAP) of pseudo labels on CIFAR10.}
	\centering
	\begin{tabular}{cccccccccc}
		\Xhline{2\arrayrulewidth}
		Metric pretraining & Propagation method & 50 & 100 & 250 & 500 & 1000 & 2000 & 4000 & 8000 \\
		\hline \hline
		\multirow{2}{*}{\makecell{Bootstrapping}} 
		&  Nearest neighbor & 22.03 & 25.74 & 48.35 & 68.03 & 77.57 & 77.28 & 87.77 & 90.88 \\
		& Spectral & 23.49 & 28.88 & 54.46 & 70.02 & 80.94 & 87.77 & \textbf{93.94} & \textbf{96.23} \\
		\hline
		
		\multirow{2}{*}{\makecell{Colorization \cite{zhang2016colorful}}}
		& Nearest neighbor & 57.32 & 67.61 & 75.48 & 79.34 & 80.70 & 82.14 & 83.66 & 84.79 \\
		& Spectral & 60.85 & 67.34 & 76.31 & 80.04 & 81.78 & 81.89 & 82.93 & 82.03 \\
		\hline
		
		\multirow{2}{*}{\makecell{Instance \cite{wu2018unsupervised}}}
		&  Nearest neighbor & 54.82 & 62.99 & 77.08 & 84.90 & 88.68 & 91.34 & 92.72 & 93.67 \\
		& Spectral & \textbf{72.59} & \textbf{79.21} & \textbf{86.64} & \textbf{90.01} & \textbf{91.04} & \textbf{91.57} & 91.77 & 91.94 \\
		\Xhline{2\arrayrulewidth}
	\end{tabular}
	\label{table:pseudo_accuracy}
\end{table*}

\begin{table*}[t]
	\setlength{\tabcolsep}{7pt}
	\caption{Ablation study of semi-supervised performance on CIFAR10.}
	\centering
	\begin{tabular}{cccccccccc}
		\Xhline{2\arrayrulewidth}
		Metric pretraining & Propagation method & 50 & 100 & 250 & 500 & 1000 & 2000 & 4000 & 8000 \\
		\hline \hline
		No & No & 20.95 & 25.35 & 41.63 & 54.06 & 65.08 & 73.22 & 81.44 & 86.23 \\
		\hline 
		
		\multirow{2}{*}{\makecell{Bootstrapping}}
		&  Nearest neighbor & 21.79 & 25.37 & 42.70 & 54.14 & 68.08 & 75.17 & 83.30 & 87.68 \\
		& Spectral & 22.78 & 27.95 & 47.28 & 60.73 & 72.60 & 78.20 & 85.10 & \textbf{88.26}  \\
		\hline
		
		\multirow{3}{*}{\makecell{Colorization \cite{zhang2016colorful}}} 
		& No & 49.57 & 55.41 & 64.65 & 68.81 & 73.40 & 77.93 & 82.17 & 86.25 \\
        & Nearest neighbor & 49.96 & 52.69 & 65.63 & 65.88 & 70.88 & 76.36 & 80.16 & 84.64	\\
		& Spectral & 53.47 & 55.08 & 68.40 & 71.15 & 72.38 & 76.50 & 80.31 & 84.03 \\
		\hline
		\multirow{3}{*}{\makecell{Instance \cite{wu2018unsupervised} }} 
		&  No & 35.27 & 37.87 & 62.46 & 71.04 & 75.96 & 80.12 & 83.90 & 87.82 \\
		&  Nearest neighbor & 46.68 & 54.45 & 66.93 & 74.16 & 79.17 & 82.24 & \textbf{84.56} & 87.92 \\
		& Spectral & \textbf{56.34} & \textbf{63.53} & \textbf{71.26} & \textbf{74.77} & \textbf{79.38} & \textbf{82.34} & 84.52 & 87.48 \\
		
		\Xhline{2\arrayrulewidth}
	\end{tabular}
	\label{table:final_accuracy}
\end{table*}

\subsection{Confidence Weighted Supervised Training}

\begin{figure}[t]
	\centering
	\includegraphics[width=0.9\linewidth]{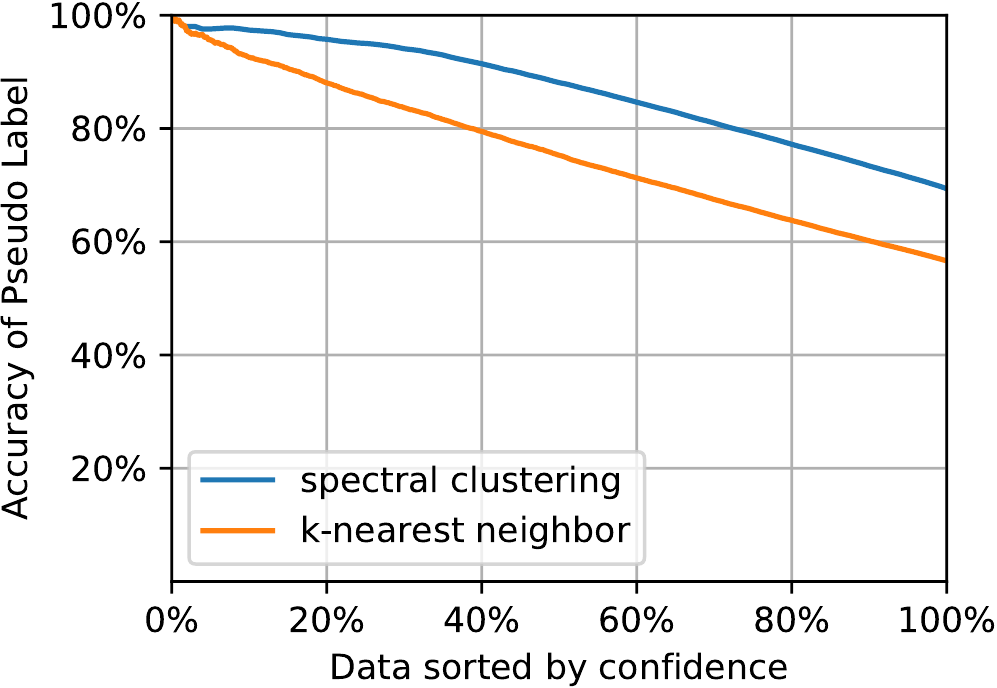}
	\caption{The accumulated accuracy of the pseudo labels
		 on the validation data sorted by the confidence measure.}
	\label{fig:confidence}
\end{figure}

Given the logits $z_i$, the pseudo label $\hat{y}_i$ is estimated as
\begin{equation}
\hat{y_i} = \argmax_c z_{i,c}.
\end{equation}
With the estimated pseudo labels on the unlabeled data,
we have considerably more data for training a classifier. 
However, the pseudo labels may not be accurate, and directly using these labels
may lead to degraded performance.
For example, not all the data in the unlabeled set are related to the target problem.
Here, we devise a simple weighting mechanism to compensate for inaccurate labels.

Given the logits $z_i$ produced by the label propagation algorithm, we first normalize it into a probabilistic distribution,
\begin{equation}
\bar{z}_{i,c} = \frac{\text{exp} (z_{i,c} / \tau)}{\sum_j \text{exp} (z_{i,j} / \tau)},
\end{equation}
where $c$ indexes the dimension of categories, and the temperature $\tau$ controls the sharpness of the distribution. We then define the confidence measure $\alpha_i$ of the pseudo label as the difference between the maximum response and the second largest response,
\begin{equation}
\alpha_i = \max_j \bar{z}_{i,j} - \max_{c\neq \argmax \bar{z}_{i,j}} \bar{z}_{i,c}.
\end{equation}
A high value of $\alpha_i$ indicates a confident estimate of the pseudo label, and a low value of $\alpha_i$ indicates an ambiguous estimate.
In Figure~\ref{fig:confidence}, we measure the accumulated accuracy of pseudo labels on validation data sorted by this confidence. It can be seen that our confidence measure gives a good indication of the quality of pseudo labels.

Our final training criterion is given by
\begin{equation}
\mathcal{L} = - \frac{1}{N} \sum_i \alpha_i \cdot \text{ log } p_{\hat{y}_i}
\end{equation}
where $\hat{y}_i$ is the pseudo label for example $i$, and $p(\cdot)$ is the softmax probability output of the classification network.

In practice, since some pseudo labels have relatively low confidence, e.g. $\alpha < 0.01$, and thus contribute negligibly to the overall learning criterion, we may safely discard those examples to speed up learning.

\section{Experiments}
Through experiments, we show that,
with unlabeled data, metric propagation is able to effectively label 
lots of data when little labeled data is given.
We verify our approach on semi-supervised learning, where an unsupervised metric is transferred, and on transfer learning, where supervised metrics generalize across different data distributions, and on few-shot recognition, where the metric can generalize across open-set object categories.
While studying few-shot recognition, we leverage an extra unlabeled data for label propagation, which is also known as semi-supervised few-shot recognition~\cite{ren2018meta}.

Our approach has two major hyper-parameters: the number of the eigenvectors $\eta$ for spectral clustering and the temperature $\sigma$ controlling the confidence distribution. Different parameter settings may slightly change the performance. We use $\eta=200$ and $\sigma=40$ across the experiments. A detailed analysis is provided in the supplementary materials.

\subsection{Semi-Supervised Learning}
\label{sec:semi}

\begin{table*}[t]
	\setlength{\tabcolsep}{8pt}
	\centering
    \caption{Scalability to large network architectures on CIFAR10.}
    \begin{tabular}{lccccccccc}
		\Xhline{2\arrayrulewidth}
	    Methods & Network architectures & 50 & 100 & 250 & 500 & 1000 & 2000 & 4000 & 8000 \\
		\hline 

        Mean Teacher & \multirow{2}{*}{\makecell{WideResNet-28-2}} & 29.66 & 36.62 & 45.49 & 57.19 & 65.07 & 79.26 & 84.38 & 87.55 \\
        \textbf{Ours} & & 56.34 & 63.53 & 71.26 & 74.77 & 79.38 & 82.34 & 84.52 & 87.48 \\
		\hline
		Mean Teacher & \multirow{2}{*}{\makecell{WideResNet-28-10}} & 27.35 & 38.83 & 49.44 & 59.45 & 70.03 & 82.62 & 86.71 & 89.38	\\
		\textbf{Ours} & & 73.13 & 75.87 & 80.30 & 81.76 & 84.97 & 86.82 & 88.70 & 91.01 \\
		
		\Xhline{2\arrayrulewidth}
	\end{tabular}
	\label{table:arch}
	\vspace{-2pt}
\end{table*}

\begin{figure}[t]
	\centering
	\includegraphics[width=0.95\linewidth]{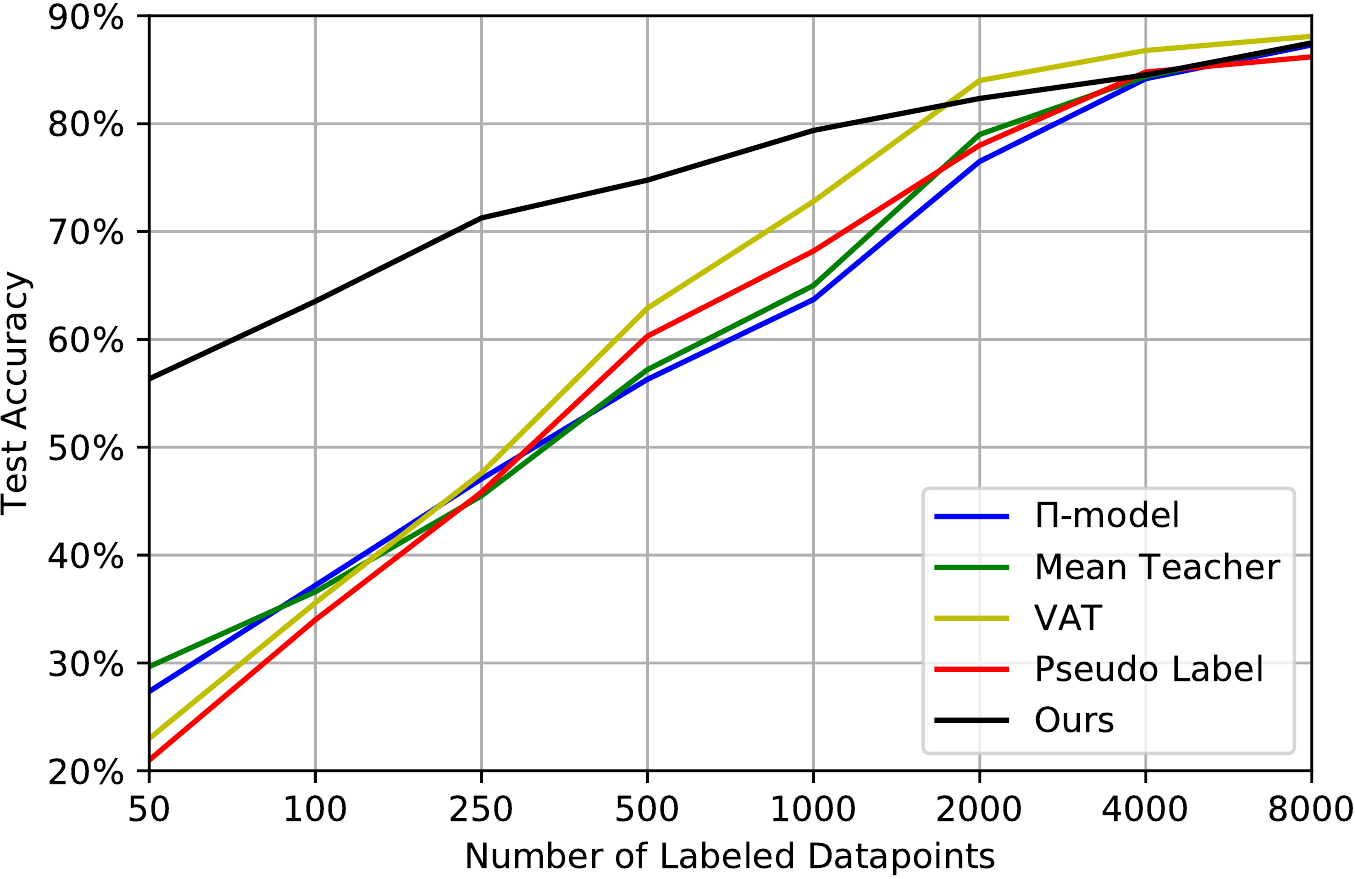}
	\caption{Comparisons to the state-of-the-art on CIFAR10.}
	\label{fig:sota}
	\vspace{-2pt}
\end{figure}

We follow a recent evaluation paper~\cite{oliver2018realistic}, which gives a comprehensive benchmark for state-of-the-art semi-supervised learning approaches.
A majority of our ablation studies are conducted on CIFAR10~\cite{krizhevsky2009learning}, while we also test our method on ImageNet.
On CIFAR10, 
we use the same Wide-ResNet~\cite{zagoruyko2016wide} architecture with 28 layers and a width factor of 2.
We report performance varying the number of labeled examples from $50$ to $8,000$ of  $50,000$ examples.

For training our model, we pretrain the metric using the unlabeled split, and propagate labels to the same unlabeled set. This means $\mathcal{S} = \mathcal{U}$ in our framework. We use SGD for optimization with an initial learning rate of 0.01 and a cosine decay schedule. We fix the total number of optimization iterations to $200K$ as opposed to fixing optimization epochs, because it gives more consistent comparisons when the number of labeled data varies.

\noindent \textbf{Study of different pretrained metrics.}

Our label propagation algorithm needs a pretrained similarity metric to guide it.
The pretrained metric can be learned by supervised methods using limited labeled data, or by unsupervised methods using large-scale unlabeled data.
Here, we consider three metric pretraining methods:
\begin{enumerate}[nolistsep]
	\setlength{\itemsep}{0pt}
	\item supervised bootstrapping on limited labeled data.
	\item self-supervised learning by image colorization~\cite{zhang2016colorful}.
	\item unsupervised learning by instance discrimination~\cite{wu2018unsupervised}.
\end{enumerate} 
We train the models using the optimal parameters for each pretraining method. 
Then we use cosine similarity in the feature space for propagating labels to the unlabeled data.

In Table~\ref{table:pseudo_accuracy}, we evaluate the quality of pseudo labels as the mean average precision (mAP) sorted by the confidence as in Figure~\ref{fig:confidence}. Table~\ref{table:final_accuracy} lists the final semi-supervised recognition accuracy. We can see that both unsupervised methods generalize much better than the supervised bootstrapping method most of the time, until the labeled set is relatively large with 4000 labels. 
This confirms our claim that unsupervised transfer is the key for label propagation.
For the unsupervised methods, non-parametric metric learning performs better than colorization, probably because it explicitly learns a similarity metric.
We also include the result of the naive baseline which trains from scratch using limited labeled data without label propagation.

\vspace{3pt}
\noindent \textbf{Study of different label propagation schemes.}

Given the pretrained metrics, there are various ways to transfer the metrics. We consider three possible solutions:
\begin{enumerate}[nolistsep]
	\setlength{\itemsep}{0pt}
	\item no propagation, only transfer network weights.
	\item nearest neighbor metric transfer.
	\item spectral metric transfer.
\end{enumerate}
The first baseline is a common practice, which basically transfers the network weights and then finetunes on the labeled data.
The second is much weaker than the third because it only considers one-hop distances, without taking into account the similarities between unlabeled pairs.

The results are summarized in Table~\ref{table:pseudo_accuracy} and Table~\ref{table:final_accuracy}.
Compared to the state-of-the-art performance in Table~\ref{table:sota_perform}, even a simple finetuning approach outperforms the state-of-the-arts when the labeled data is small. For example, by finetuning from instance discrimination, we achieve $62.46\%$ with $250$ labeled data, significantly outperforming the state-of-the-art result of $47.07\%$.
This suggests that unsupervised pretraining generally improves semi-supervised learning.

When unlabeled data is used for label propagation, 
metric transfer can be much stronger than just weight transfer, improving the performance to $71.26\%$ with $250$ labeled data.
It is also evident that the spectral clustering method performs better
than weighted nearest neighbors because of its globalization behavior.
\begin{table}[t]
    \setlength{\tabcolsep}{12pt}
    \centering
    \caption{Ours is complementary to all prior state-of-the-art methods on CIFAR10.}
    \begin{tabular}{lcc}
    \Xhline{2\arrayrulewidth}
    Num Labeled & 250 & 4000 \\
    \hline \hline
    Ours &  71.26 &	84.52 \\
   	\hline
    Pi Model~\cite{laine2016temporal} & 47.07 & 84.17 \\ 
    \textbf{+ Ours} & 74.90 & 85.32 \\
    \hline
    Mean Teacher~\cite{tarvainen2017mean} & 45.49 & 84.38 \\
    \textbf{+ Ours} & 74.54 & 85.45 \\
    \hline
    VAT~\cite{miyato2017virtual} & 44.83 & 86.79\\
    \textbf{+ Ours} & 78.34 & 86.93 \\
	\hline
    VAT+EM~\cite{miyato2017virtual} & 46.29 & 86.96\\
    \textbf{+ Ours} & 78.63 &  87.20 \\
    \Xhline{2\arrayrulewidth}
    \end{tabular}
    \label{table:sota_perform}
    	\vspace{-2pt}
\end{table}

\noindent \textbf{Scalability to large network architectures.}

In contrast to prior methods which face over-fitting issues, our approach can easily scale to larger network architectures. 
Here, we keep all the learning parameters unchanged, and experiment with a wider version of Wide-ResNet-28 with a width factor of 10.
We consider a state-of-the-art method mean-teacher~\cite{tarvainen2017mean} for comparison.
In Table~\ref{table:arch}, mean-teacher only shows a limited improvement of about $2-3\%$.
Our method enjoys consistently significant gains from a larger network on all the testing scenarios. It achieves an unprecedented $73.13\%$ accuracy using only $50$ labels with Wide-ResNet-28-10.

\vspace{3pt}

\noindent \textbf{Comparison to the state-of-the-art on CIFAR10.}

We compare our approach to state of the art methods in Figure~\ref{fig:sota}.
Ours is particularly stronger when the labeled set is small, but 
this advantage diminishes as the labeled set grows.
However, as most prior approaches focus on self-ensembling, ours is orthogonal to them.
We examine the complementarity of our method by combining it with each of the prior approaches.
To do so, we generate our most confident $10K$ pseudo labels (about $20\%$ of the full data), and use it as ground-truth for the other algorithms.
For fair comparisons, we run public code\footnote{https://github.com/brain-research/realistic-ssl-evaluation} with our generated pseudo labels.
In Table~\ref{table:sota_perform},
combining our approach leads to improved performance for all of the methods.

\vspace{3pt}

\noindent \textbf{Comparison to the state-of-the-art on ImageNet.}

We notice that few literature report semi-supervised classification performance on ImageNet consistently.
In this paper, we consider finetuning from an unsupervised model trained with instance discrimination~\cite{wu2018unsupervised} as our baseline.
We vary the number of labeled examples from $1\%$ to $4\%$ of the entire ImageNet.
We use ResNet-50 to pretrain the unsupervised model, and split the dataset into 10 chunks for spectral clustering to speed up the computation.
In Table~\ref{table:imagenet_semi}, finetuning from unsupervised model significantly improves upon training from scratch, and our label propagation approach outperforms the finetuning approach. Notable, ours is $18\%$ better when $1\%$ labeled data is available.

\begin{table}[t]
    \setlength{\tabcolsep}{12pt}
    \centering
    \caption{Semi-supervised classification results on the ImageNet dataset.}
    \begin{tabular}{lccc}
    \Xhline{2\arrayrulewidth}
    Num Labeled & $1\%$ & $2\%$ & $4\%$\\
    \hline
    Scratch & 22.4 & 40.2 & 58.2\\
    Finetune & 39.2 & 52.8 & 65.2 \\
    \textbf{Ours} & \textbf{58.6} & \textbf{66.3} & \textbf{72.4} \\
    \Xhline{2\arrayrulewidth}
    \end{tabular}
    \label{table:imagenet_semi}
    \vspace{-2pt}
\end{table}

\subsection{Transfer Learning}

We also examine whether the proposed metric transfer can work across different data distributions. We pretrain the metric on the source $\mathcal{S}$ ImageNet, and transfer it to the unlabeled $\mathcal{U}$ CIFAR10.
For this, we study supervised and unsupervised pretraining for transfer learning.

\begin{table*}[t]
    \caption{Transfer learning from ImageNet to CIFAR10.}
	\setlength{\tabcolsep}{8pt}
	\centering
	\begin{tabular}{cccccccccc}
		\Xhline{2\arrayrulewidth}
		Metric pretraining & Transfer method & 50 & 100 & 250 & 500 & 1000 & 2000 & 4000 & 8000 \\
		\hline \hline
        \multirow{2}{*}{\makecell{Unsupervised}} & 
		Network finetuning & 28.92 & 34.56 & 57.14 & 67.54 & 76.20 & 80.92 & 85.01 & 88.74 \\
		& Spectral & 44.30 & 46.51 & 61.29 & 68.31 & 72.61 & 77.86 & 84.00 & 88.19 \\
		\hline
        \multirow{2}{*}{\makecell{Supervised}} & 
        Network finetuning & 54.95 & 61.88 & 73.01 & 78.43 & 84.52 & 88.79 & 91.44 & 93.05 \\
        & Spectral & 77.71 & 85.34 & 86.07 & 86.91 & 88.27 & 89.93 & 91.22 & 93.49 \\
		\Xhline{2\arrayrulewidth}
	\end{tabular}
	\label{table:transfer}
\end{table*}

\vspace{3pt}
\noindent \textbf{Transferring from labeled ImageNet.}
We resize ImageNet images to a resolution of $32\times 32$ and pretrain
the metric on them by supervised learning.
We keep the network architecture WideResNet-28-2 for meaningful comparison with the semi-supervised settings in Sec~\ref{sec:semi}.
This obtains an accuracy of $42\%$ on the ImageNet validation set.
Then we transfer the metric to CIFAR10.
This transfer is conducted by network finetuning and by metric propagation.
In Table~\ref{table:transfer}, we can see that simple network finetuning can reach the best results obtained in the semi-supervised settings of the previous subsection.
By using label propagation with spectral clustering, 
we can observe a large improvement, yielding $86.07\%$ accuracy with just $250$ labeled images.
This illustrates the generality of our metric transfer approach, where supervised transfer can also
take advantage of unlabeled data to improve generalization.

\vspace{3pt}
\noindent \textbf{Transferring from unlabeled ImageNet.}
Instead of supervised training which encodes prior knowledge about object categories,
we treat ImageNet images as unlabeled and repeat the previous experiment.
Different from the earlier unsupervised experiments, this setting involves substantially more unlabeled data, which could potentially lead to a better unsupervised metric.
However, our results suggest otherwise.
When propagating to CIFAR10, the unsupervised metric learned from ImageNet is inferior to the metric learned from CIFAR10.
This is possibly due to the data distribution gap between CIFAR10 and ImageNet.
Nevertheless, our unsupervised transfer from ImageNet still surpasses
the state-of-the-art in the semi-supervised setting when labeled samples are limited.

\subsection{Few-Shot Recognition}

\begin{table}[t]
	\setlength{\tabcolsep}{3.2pt}
	\centering
	\caption{Few-shot recognition on Mini-ImageNet dataset.}
	\begin{tabular}{lcccc}
		\Xhline{2\arrayrulewidth} 
		\multirow{2}{*}{Method} &  \multirow{2}{*}{Fintune} &  Unlabel & \multicolumn{2}{c}{5-way Classification}  \\
		&  & data & 1-shot & 5-shot \\
		\hline \hline
		NN baseline~\cite{vinyals2016matching}  & No & No & 41.1$\pm$0.7 & 51.0$\pm$0.7 \\
		MAML~\cite{finn2017model} & Yes & No & 48.7$\pm$0.7 & 63.2$\pm$0.9 \\
		Meta-SGD~\cite{li2017meta} & No & No & 50.5$\pm$1.9 & 64.0$\pm$0.9 \\
		Matching net~\cite{vinyals2016matching} & Yes & No & 46.6$\pm$0.8  & 60.0$\pm$0.7   \\
		Prototypical~\cite{snell2017prototypical} & No & No & 49.4$\pm$0.8 & 68.2$\pm$0.7\\
	
		SNCA~\cite{wu2018improving} & No & No & 50.3$\pm$0.7 & 64.1$\pm$0.8 \\
		\hline
			Soft k-means~\cite{ren2018meta} & Yes & Yes & 50.4$\pm$0.3 & 64.4$\pm$0.2\\
			
		Our supervised & Yes & Yes & \textbf{56.1}$\pm$0.6 & \textbf{70.7}$\pm$0.5\\
		Our unsupervised & Yes & Yes & 50.8$\pm$0.6&  66.0$\pm$0.5\\
		\Xhline{2\arrayrulewidth}
	\end{tabular}
		\vspace{-2pt}
	\label{table:few-shot}
\end{table}

Few-shot recognition targets a more challenging scenario, the generalization across object categories (\textit{a.k.a.} open-set recognition).
Originally, the problem is defined with numerous labeled examples in a source dataset, and few examples in the target categories.
Recent works~\cite{ren2018meta,garcia2017few} also explore the scenario where extra unlabeled data is available for this problem.
This fits into our framework for studying label propagation via metric transfer.

We follow the protocols in~\cite{ren2018meta} for conducting the experiments, because it introduces distractor categories in the unlabeled set.
The experiments are evaluated on the mini-ImageNet dataset, consisting of a total of $100$ categories, with $64$ for training, $16$ for validation and $20$ for testing.
Images in each category are split into $40\%$ as labeled, and $60\%$ as unlabeled.
Training uses only the labeled split in the training categories.
During evaluation, 
a testing episode is constructed by sampling few-shot labeled observations from the labeled split in the testing categories, and all of the unlabeled images in all the testing categories.
A testing episode requires the model to find useful information in the unlabeled set to aid recognition from the few-shot observations.
Unlike~\cite{ren2018meta}, which includes five distractor categories in the unlabeled set, 
we consider all $20$ categories in the testing set, which better reflects practical scenarios.
We test $300$ episodes and report the results.

We follow prior work~\cite{vinyals2016matching} by using a shallow architecture with four convolutional layers and a final fully connected layer. Each convolutional layer has 64 channels, interleaved with ReLU, subsampling and a batch normalization layer.
Images are resized to $84\times 84$ to train the model.
We use the spectral embedding approach for label propagation.
During online training, we use an initial learning rate of $0.01$ with a total of 30 epochs and decrease the learning rate to be $5$ times smaller after $20$ epochs.

\begin{figure}[t]
	\centering
	\includegraphics[width=0.9\linewidth]{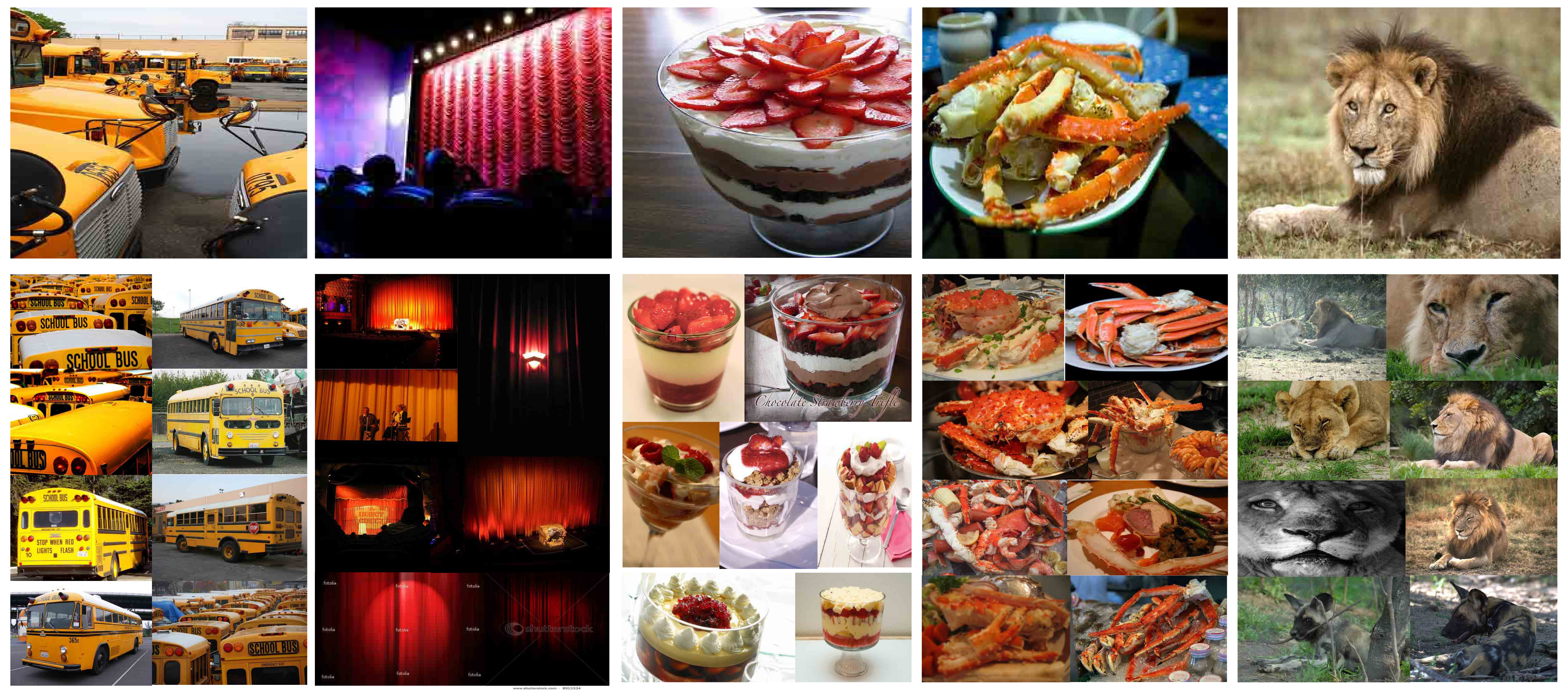}
	\caption{Visualizations of top ranked retrievals from the unlabeled set given one-shot observations.}
	\label{fig:fewshot}
	\vspace{-2pt}
\end{figure}

\vspace{3pt}
\noindent \textbf{Transfer from supervised models.}
We use a recent supervised metric learning approach SNCA~\cite{wu2018improving} as the baseline.
After label propagation and finetuning on the new data, our supervised propagation obtains a significant boost of $6\%$ over SNCA.
Prior work~\cite{ren2018meta} improves upon its baselines, but fails to make further improvement because of limited training data.
In Figure~\ref{fig:fewshot}, we visualize the top retrievals from the unlabeled set in the one-shot scenario. These retrievals not only belong to the same class as the ground truth, but their diversity facilitates a strong classifier.

\vspace{3pt}
\noindent \textbf{Transfer from unsupervised models.}
We also investigate pretraining the metric without labels, using instance discrimination~\cite{wu2018unsupervised} for learning the metric.
Surprisingly, in Table~\ref{table:few-shot}, our unsupervised propagation obtains better performance than the offline metric learning approach with annotations~\cite{wu2018improving}, by $0.5\%$ in 1-shot recognition and $2\%$ for 5-shot.
This suggests that leveraging unlabeled data in the target problem may possibly be more beneficial than using labeled samples in the source domain.

\begin{figure*}[t]
	\centering
	\includegraphics[width=0.45\linewidth]{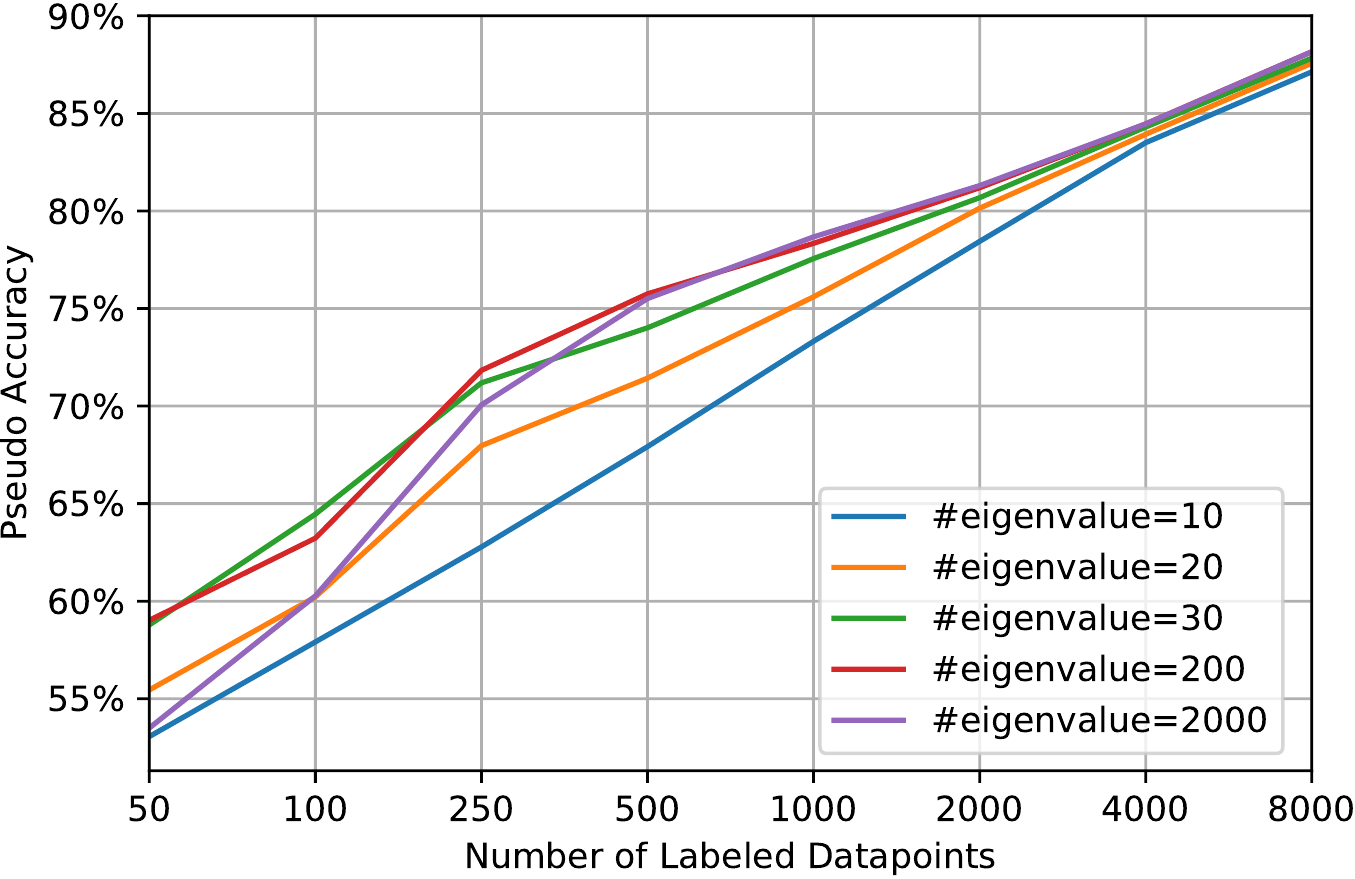}
	\includegraphics[width=0.45\linewidth]{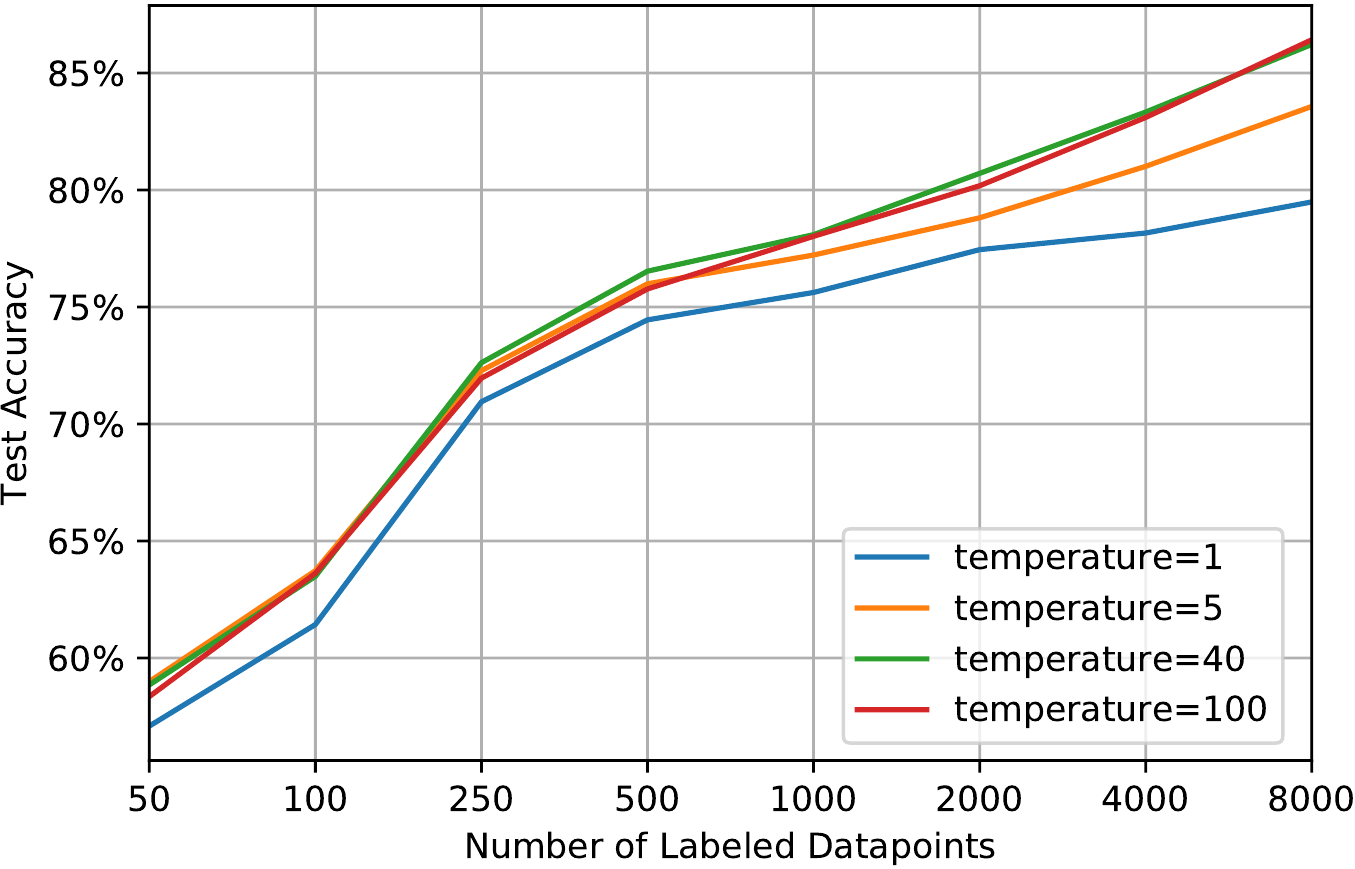}
	\caption{Ablations of model parameters $\eta$ and $\sigma$.}
	\label{fig:parameters}
\end{figure*}

\section{Discussions}


\begin{itemize}[nolistsep]

    \item The effectiveness of label propagation depends heavily on the learned metric, so advances in metric learning should lead to improved results.
    Since the prevalent pretraining methods in deep learning use softmax classification, we hope to draw more attention to pretraining networks with metric learning.
    
    \item Currently, we study metric pretraining and label propagation separately. It may be beneficial to formulate them jointly in an end-to-end framework.
    
    
    \item Our algorithm takes advantage of the
    unlabeled dataset $\mathcal{U}$ to create more training data. 
    The overall performance is affected by the relevance of image content in the unlabeled set $\mathcal{U}$ to that of the target $\mathcal{T}$, as this impacts the ability to effectively propagate labels.
    
\end{itemize}

\appendix
\renewcommand{\thesection}{A\arabic{section}}

\begin{figure*}[t]
	\centering
	\includegraphics[width=0.92\linewidth]{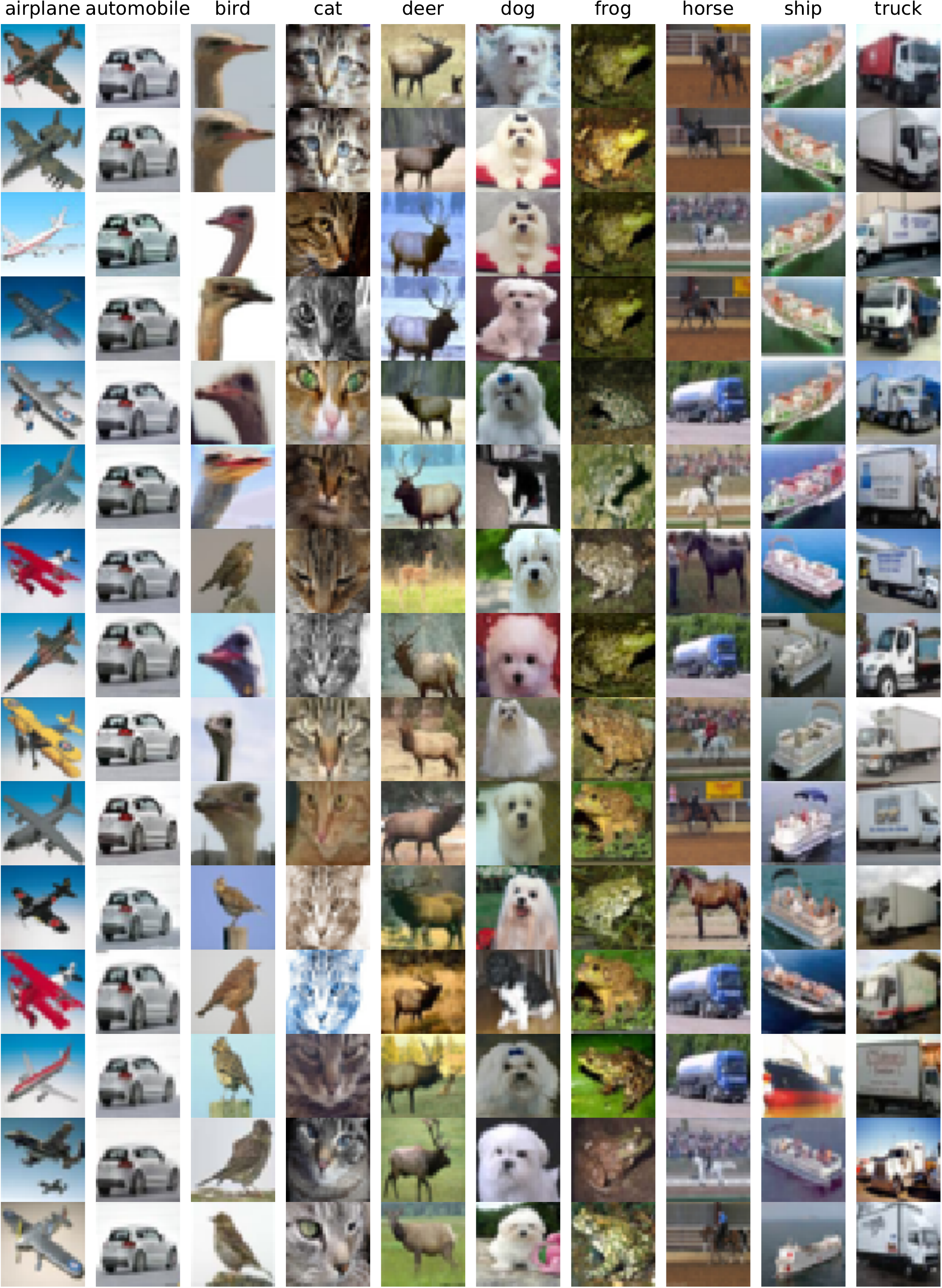}
	\caption{CIFAR10 retrievals for 10 categories.}
	\label{fig:cifar10}
\end{figure*}

\begin{figure*}[t]
	\centering
	\includegraphics[width=0.9\linewidth]{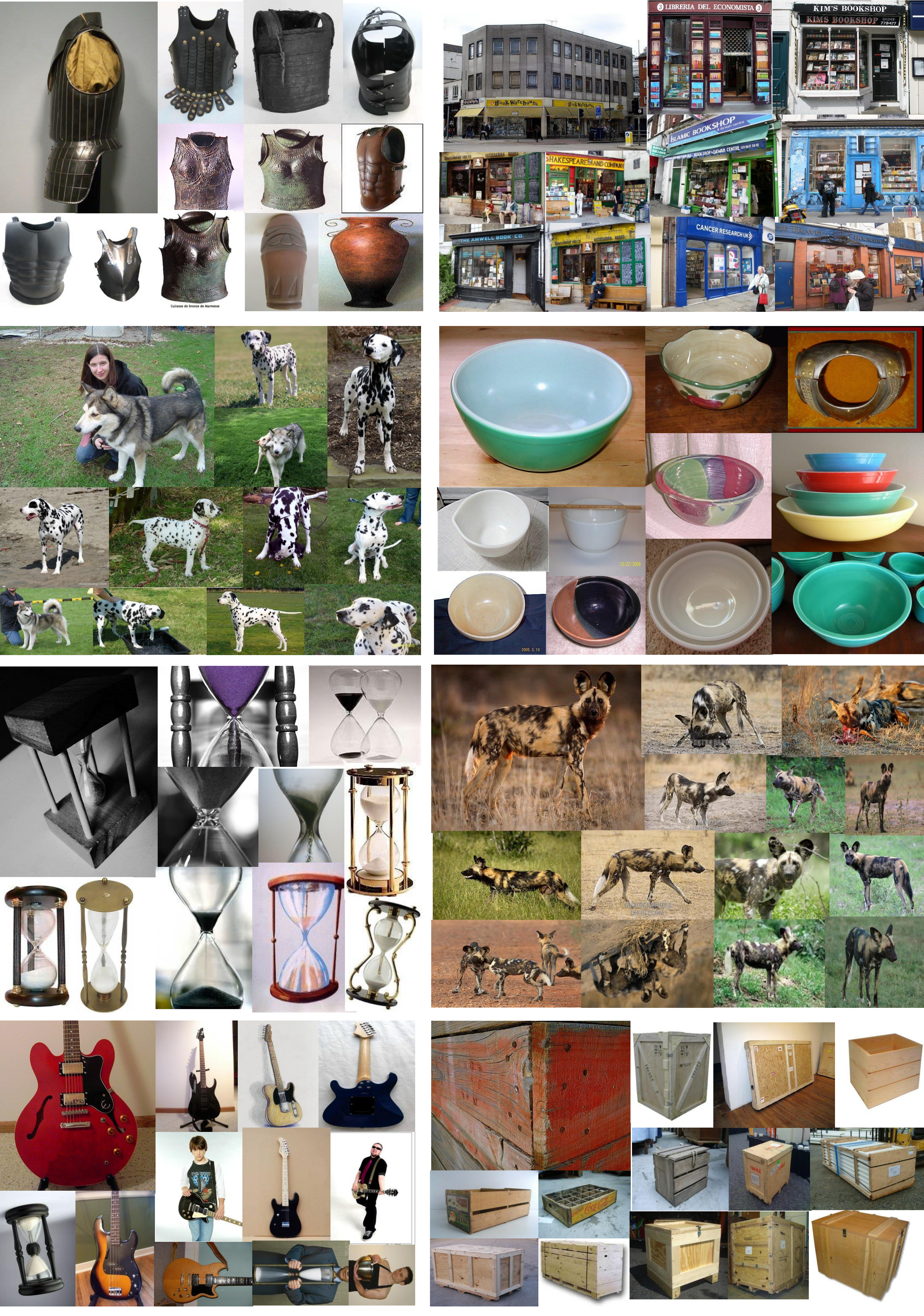}
	\caption{Mini-ImageNet retrieval visualizations. Top left is the one-shot query, the rest are the top retrievals.}
	\label{fig:mini-imagenet}
\end{figure*}

\section{Ablations of Model Parameters}
Our model depends on two parameters: the number of eigen components $\eta$ used for spectral clustering, and the temperature $\sigma$ used for controlling the confidence.  We used $\eta=200$ and $\sigma=40$ in our main submission. In Figure~\ref{fig:parameters}, we show the effects of the two parameters respectively.

The number of eigenvectors $\eta$ works well in the range between $30$ and $200$. We can see a trade-off of the value $\eta$ for performance under various number of labeled samples. Smaller $\eta$ benefits very few labeled samples, while larger $\eta$ benefits comparably more labeled samples.
For the temperature parameter $\sigma$, it is generally robust for a wide range of values between $10$ to $100$.

\section{Additional Visualizations}
We provide more retrieval visualizations in the CIFAR10 and mini-ImageNet dataset in Figure~\ref{fig:cifar10} and Figure~\ref{fig:mini-imagenet}. For CIFAR10, we show the top retrievals for each class in the unlabeled set given $250$ labeled examples. For mini-ImageNet, we show the top retrievals in the 5-class 1-shot scenario.

{\small
\bibliographystyle{ieee}
\bibliography{egbib}

\begin{thebibliography}{10}\itemsep=-1pt

\bibitem{carey1978acquiring}
S.~Carey and E.~Bartlett.
\newblock Acquiring a single new word.
\newblock 1978.

\bibitem{clark_cross-view_2018}
K.~Clark, T.~Luong, and Q.~V. Le.
\newblock Cross-view training for semi-supervised learning.
\newblock In {\em ICLR}, 2019.

\bibitem{deng2009imagenet}
J.~Deng, W.~Dong, R.~Socher, L.-J. Li, K.~Li, and L.~Fei-Fei.
\newblock Imagenet: A large-scale hierarchical image database.
\newblock In {\em CVPR}. Ieee, 2009.

\bibitem{doersch2015unsupervised}
C.~Doersch, A.~Gupta, and A.~A. Efros.
\newblock Unsupervised visual representation learning by context prediction.
\newblock In {\em ICCV}, 2015.

\bibitem{doersch2017multi}
C.~Doersch and A.~Zisserman.
\newblock Multi-task self-supervised visual learning.
\newblock In {\em ICCV}, 2017.

\bibitem{everingham2010pascal}
M.~Everingham, L.~Van~Gool, C.~K. Williams, J.~Winn, and A.~Zisserman.
\newblock The pascal visual object classes (voc) challenge.
\newblock {\em IJCV}, 2010.

\bibitem{fergus2009semi}
R.~Fergus, Y.~Weiss, and A.~Torralba.
\newblock Semi-supervised learning in gigantic image collections.
\newblock In {\em NIPS}, 2009.

\bibitem{finn2017model}
C.~Finn, P.~Abbeel, and S.~Levine.
\newblock Model-agnostic meta-learning for fast adaptation of deep networks.
\newblock {\em arXiv preprint arXiv:1703.03400}, 2017.

\bibitem{garcia2017few}
V.~Garcia and J.~Bruna.
\newblock Few-shot learning with graph neural networks.
\newblock {\em arXiv preprint arXiv:1711.04043}, 2017.

\bibitem{gastaldi2017shake}
X.~Gastaldi.
\newblock Shake-shake regularization.
\newblock {\em arXiv preprint arXiv:1705.07485}, 2017.

\bibitem{girshick2014rich}
R.~Girshick, J.~Donahue, T.~Darrell, and J.~Malik.
\newblock Rich feature hierarchies for accurate object detection and semantic
  segmentation.
\newblock In {\em CVPR}, 2014.

\bibitem{goldberger2005neighbourhood}
J.~Goldberger, G.~E. Hinton, S.~T. Roweis, and R.~R. Salakhutdinov.
\newblock Neighbourhood components analysis.
\newblock In {\em Advances in neural information processing systems}, 2005.

\bibitem{hoffer2015deep}
E.~Hoffer and N.~Ailon.
\newblock Deep metric learning using triplet network.
\newblock In {\em International Workshop on Similarity-Based Pattern
  Recognition}. Springer, 2015.

\bibitem{hoffman2017cycada}
J.~Hoffman, E.~Tzeng, T.~Park, J.-Y. Zhu, P.~Isola, K.~Saenko, A.~A. Efros, and
  T.~Darrell.
\newblock Cycada: Cycle-consistent adversarial domain adaptation.
\newblock {\em arXiv preprint arXiv:1711.03213}, 2017.

\bibitem{hu2013multi}
H.~Hu, J.~Feng, C.~Yu, and J.~Zhou.
\newblock Multi-class constrained normalized cut with hard, soft, unary and
  pairwise priors and its applications to object segmentation.
\newblock {\em TIP}, 2013.

\bibitem{joachims2003transductive}
T.~Joachims.
\newblock Transductive learning via spectral graph partitioning.
\newblock In {\em Proceedings of the 20th International Conference on Machine
  Learning (ICML-03)}, pages 290--297, 2003.

\bibitem{kingma2014semi}
D.~P. Kingma, S.~Mohamed, D.~J. Rezende, and M.~Welling.
\newblock Semi-supervised learning with deep generative models.
\newblock In {\em NIPS}, 2014.

\bibitem{krizhevsky2009learning}
A.~Krizhevsky and G.~Hinton.
\newblock Learning multiple layers of features from tiny images.
\newblock Technical report, Citeseer, 2009.

\bibitem{krizhevsky2012imagenet}
A.~Krizhevsky, I.~Sutskever, and G.~E. Hinton.
\newblock Imagenet classification with deep convolutional neural networks.
\newblock In {\em Advances in neural information processing systems}, 2012.

\bibitem{laine2016temporal}
S.~Laine and T.~Aila.
\newblock Temporal ensembling for semi-supervised learning.
\newblock {\em arXiv preprint arXiv:1610.02242}, 2016.

\bibitem{lake2011one}
B.~Lake, R.~Salakhutdinov, J.~Gross, and J.~Tenenbaum.
\newblock One shot learning of simple visual concepts.
\newblock In {\em Proceedings of the Annual Meeting of the Cognitive Science
  Society}, 2011.

\bibitem{lee2013pseudo}
D.-H. Lee.
\newblock Pseudo-label: The simple and efficient semi-supervised learning
  method for deep neural networks.
\newblock In {\em Workshop on Challenges in Representation Learning, ICML},
  2013.

\bibitem{li2017meta}
Z.~Li, F.~Zhou, F.~Chen, and H.~Li.
\newblock Meta-sgd: Learning to learn quickly for few shot learning.
\newblock {\em arXiv preprint arXiv:1707.09835}, 2017.

\bibitem{lin2014microsoft}
T.-Y. Lin, M.~Maire, S.~Belongie, J.~Hays, P.~Perona, D.~Ramanan,
  P.~Doll{\'a}r, and C.~L. Zitnick.
\newblock Microsoft coco: Common objects in context.
\newblock In {\em ECCV}. Springer, 2014.

\bibitem{long2015fully}
J.~Long, E.~Shelhamer, and T.~Darrell.
\newblock Fully convolutional networks for semantic segmentation.
\newblock In {\em CVPR}, pages 3431--3440, 2015.

\bibitem{mishra2018simple}
N.~Mishra, M.~Rohaninejad, X.~Chen, and P.~Abbeel.
\newblock A simple neural attentive meta-learner.
\newblock 2018.

\bibitem{miyato2017virtual}
T.~Miyato, S.-i. Maeda, M.~Koyama, and S.~Ishii.
\newblock Virtual adversarial training: a regularization method for supervised
  and semi-supervised learning.
\newblock 2017.

\bibitem{oliver2018realistic}
A.~Oliver, A.~Odena, C.~Raffel, E.~D. Cubuk, and I.~J. Goodfellow.
\newblock Realistic evaluation of semi-supervised learning algorithms.
\newblock 2018.

\bibitem{rasmus2015semi}
A.~Rasmus, M.~Berglund, M.~Honkala, H.~Valpola, and T.~Raiko.
\newblock Semi-supervised learning with ladder networks.
\newblock In {\em NIPS}, 2015.

\bibitem{ren2018meta}
M.~Ren, E.~Triantafillou, S.~Ravi, J.~Snell, K.~Swersky, J.~B. Tenenbaum,
  H.~Larochelle, and R.~S. Zemel.
\newblock Meta-learning for semi-supervised few-shot classification.
\newblock {\em arXiv preprint arXiv:1803.00676}, 2018.

\bibitem{rosenberg2005semi}
C.~Rosenberg, M.~Hebert, and H.~Schneiderman.
\newblock Semi-supervised self-training of object detection models.
\newblock In {\em WACV/MOTION}, 2005.

\bibitem{russakovsky2015imagenet}
O.~Russakovsky, J.~Deng, H.~Su, J.~Krause, S.~Satheesh, S.~Ma, Z.~Huang,
  A.~Karpathy, A.~Khosla, M.~Bernstein, et~al.
\newblock Imagenet large scale visual recognition challenge.
\newblock {\em IJCV}, 2015.

\bibitem{shi2000normalized}
J.~Shi and J.~Malik.
\newblock Normalized cuts and image segmentation.
\newblock {\em TPAMI}, 2000.

\bibitem{simonyan2014very}
K.~Simonyan and A.~Zisserman.
\newblock Very deep convolutional networks for large-scale image recognition.
\newblock {\em arXiv preprint arXiv:1409.1556}, 2014.

\bibitem{snell2017prototypical}
J.~Snell, K.~Swersky, and R.~Zemel.
\newblock Prototypical networks for few-shot learning.
\newblock In {\em NIPS}, 2017.

\bibitem{tarvainen2017mean}
A.~Tarvainen and H.~Valpola.
\newblock Mean teachers are better role models: Weight-averaged consistency
  targets improve semi-supervised deep learning results.
\newblock In {\em NIPS}, 2017.

\bibitem{vinyals2016matching}
O.~Vinyals, C.~Blundell, T.~Lillicrap, D.~Wierstra, et~al.
\newblock Matching networks for one shot learning.
\newblock In {\em NIPS}, 2016.

\bibitem{von2007tutorial}
U.~Von~Luxburg.
\newblock A tutorial on spectral clustering.
\newblock {\em Statistics and computing}, 2007.

\bibitem{weston2012deep}
J.~Weston, F.~Ratle, H.~Mobahi, and R.~Collobert.
\newblock Deep learning via semi-supervised embedding.
\newblock In {\em Neural Networks: Tricks of the Trade}. Springer, 2012.

\bibitem{wu2018improving}
Z.~Wu, A.~A. Efros, and S.~X. Yu.
\newblock Improving generalization via scalable neighborhood component
  analysis.
\newblock {\em arXiv preprint arXiv:1808.04699}, 2018.

\bibitem{wu2018unsupervised}
Z.~Wu, Y.~Xiong, X.~Y. Stella, and D.~Lin.
\newblock Unsupervised feature learning via non-parametric instance
  discrimination.
\newblock In {\em CVPR}, 2018.

\bibitem{xu2017unified}
Y.~Xu, S.~J. Pan, H.~Xiong, Q.~Wu, R.~Luo, H.~Min, and H.~Song.
\newblock A unified framework for metric transfer learning.
\newblock {\em IEEE Trans. Knowl. Data Eng.}, 2017.

\bibitem{zagoruyko2016wide}
S.~Zagoruyko and N.~Komodakis.
\newblock Wide residual networks.
\newblock {\em arXiv preprint arXiv:1605.07146}, 2016.

\bibitem{zhang2016colorful}
R.~Zhang, P.~Isola, and A.~A. Efros.
\newblock Colorful image colorization.
\newblock In {\em ECCV}. Springer, 2016.

\bibitem{zhou2004learning}
D.~Zhou, O.~Bousquet, T.~N. Lal, J.~Weston, and B.~Sch{\"o}lkopf.
\newblock Learning with local and global consistency.
\newblock In {\em Advances in neural information processing systems}, pages
  321--328, 2004.

\bibitem{zhou2017unsupervised}
T.~Zhou, M.~Brown, N.~Snavely, and D.~G. Lowe.
\newblock Unsupervised learning of depth and ego-motion from video.
\newblock In {\em CVPR}, 2017.

\end{thebibliography}
}

\end{document}